\documentclass[10pt,twocolumn,letterpaper]{article}

\usepackage{wacv}              % To produce the CAMERA-READY version
\usepackage{graphicx}
\usepackage{amsmath}
\usepackage{amssymb}
\usepackage{booktabs}
\usepackage{overpic}
\usepackage{tabularx}
\usepackage{enumitem}
\usepackage[accsupp]{axessibility}

\usepackage[pagebackref,breaklinks,colorlinks]{hyperref}

\usepackage{orcidlink}
\newcommand{\ra}[1]{\renewcommand{\arraystretch}{#1}}
\usepackage{mathtools}
\usepackage{xspace}

\newcommand{\methodname}{GANFusion\xspace}
\usepackage[capitalize]{cleveref}
\crefname{section}{Sec.}{Secs.}
\Crefname{section}{Section}{Sections}
\Crefname{table}{Table}{Tables}
\crefname{table}{Tab.}{Tabs.}

\def\wacvPaperID{883} % *** Enter the WACV Paper ID here
\def\confName{WACV}
\def\confYear{2025}

\begin{document}

\title{\methodname: Feed-Forward Text-to-3D with Diffusion in GAN Space} 

% \and

\author{
    Souhaib Attaiki$^{1, *}$,
    Paul Guerrero$^{2}$,
    Duygu Ceylan$^{2}$,
    Niloy J. Mitra$^{2,3}$,
    Maks Ovsjanikov$^{1}$ \\[1ex]
    $^1$\small{LIX, \'{E}cole Polytechnique, IPP Paris}, 
    $^2$\small{Adobe Research}, 
    $^3$\small{University College London (UCL)}
    \\
    $^{*}$ \small{Work done at Adobe Research}\\\\
    \url{https://ganfusion.github.io/}
}

\maketitle

\begin{abstract}

We train a feed-forward text-to-3D diffusion generator for human characters using only single-view 2D data for supervision. 
Existing 3D generative models cannot yet match the fidelity of image and/or video generative models. 
State-of-the-art 3D generators are either trained with explicit 3D supervision and are thus limited by the volume and diversity of existing 3D data. Meanwhile, generators that can be trained with only 2D data as supervision typically produce coarser results, cannot be text-conditioned, and/or must revert to test-time optimization. 
We observe that GAN- and diffusion-based generators have complementary qualities: GANs can be trained efficiently with 2D supervision to produce high-quality 3D objects but are hard to condition on text. In contrast, denoising diffusion models can be conditioned efficiently but tend to be hard to train with only 2D supervision.
We introduce \methodname that starts by generating unconditional triplane features for 3D data using a GAN architecture trained with only \textit{single-view 2D data.} We then generate random samples from the GAN, caption them, and train a text-conditioned diffusion model that directly learns to sample from the space of good triplane features that can be decoded into 3D objects. 
We evaluate the proposed method in the context of text-conditioned full-body human generation and show improvements over possible alternatives. %In summary, we train and utilize a GAN to obtain a latent triplane space that subsequently enables, for the first time, training a text-guided feed-forward 3D generative model with only easy-to-obtain 2D data. 
%\newtext{Our code will be made available upon acceptance.} 

\if0
While text-to-image generation has shown great results using relatively mature diffusion models, the design space for text-to-3D generation is still being explored. Current text-to-3D generators suffer from different types of drawbacks: methods supervised with 3D data have poor quality or generality, due to the limited availability of 3D data; methods supervised with single-view 2D data either have limited output quality, or require lengthy optimization in the order of tens of minutes per scene.

We propose a feed-forward text-to-3D generator that can be trained with only single-view 2D images as supervision.
We observe that GANs can be trained efficiently with 2D supervision to get high-quality 3D scenes, but are hard to condition on text, while diffusion models can be conditioned efficiently on text, but are hard to train with only 2D supervision.
%Our idea is to train a diffusion model on a latent space of triplanes that was learned by a GAN with 2D supervision. The GAN gives us high-quality 3D scenes from 2D supervision, but is hard to condition on text. The diffusion model can be conditioned efficiently on text, but is hard to train with only 2D supervision.
Our idea is to combine the advantages of both methods by generating a large set of unconditional 3D scenes with a GAN, that can then be captioned and used as 3D supervision to train a text-to-3D diffusion model.
\fi

\end{abstract}

% \vspace{-2em}

\section{Introduction}
% \vspace{-1em}
Text-to-image diffusion models that operate in pixel~\cite{ho2020denoising} or latent spaces~\cite{Rombach_2022_CVPR} are still relatively recent developments, but the underlying architectures have been studied extensively, resulting in multiple commercial systems across the industry that have given rise to new creative workflows.

A similar success story is yet to be repeated for 3D data. Diffusion-based methods are known to require significant amounts of training data in order to produce high-quality generative results. Unfortunately obtaining such amounts of data in 3D is still challenging. Thus relatively few such methods have been developed with explicit 3D supervision (e.g.~\cite{Wang2022RODINAG,instant3d,zeng2022lion,hyperdiffusion} among others). 

An attractive alternative is to train 3D generative models with easier-to-obtain 2D data. In this context, one possibility is to use 2D images as a supervisory signal for a 3D generative model~\cite{anciukevivcius2023renderdiffusion}. Unfortunately, methods that use this approach tend to generate 3D models with lower quality and/or diversity. A different option is to produce high-quality 3D output by distilling pretrained 2D image diffusion priors~\cite{poole2022dreamfusion}. However, such methods require costly optimization, e.g., using Score Distillation Sampling, at inference time, making them 
%quite
slow and difficult to scale.
\begin{figure*}[!ht]
  \includegraphics[width=\textwidth]{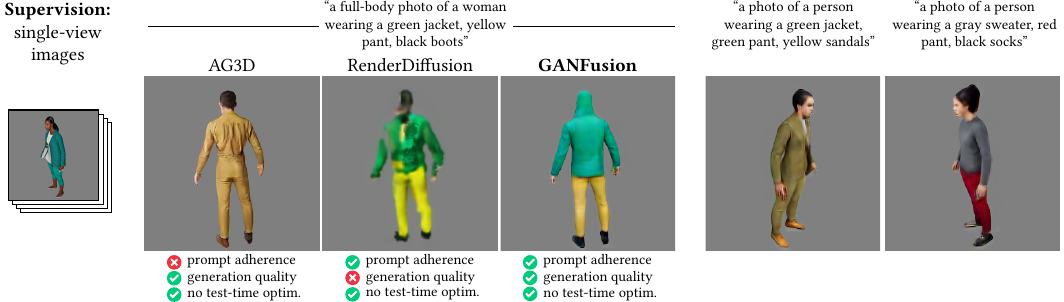}
  \caption{We propose \methodname, a text-guided feed-forward 3D generator that is trained with only single-view image supervision. Unlike previous methods such as AG3D~\cite{dong2023ag3d} which do not enable text conditioning, \methodname can be conditioned on text while still achieving high generation quality compared to other text conditioned generators such as RenderDiffusion~\cite{anciukevivcius2023renderdiffusion}. We note that \methodname, unlike SDS-based optimization methods that use image diffusion priors~\cite{poole2022dreamfusion}, does not require any test-time optimization.}
  \label{fig:teaser}
  \vspace{-1.0em}
\end{figure*}
On the other hand, GAN-based methods have shown to be possible to train with only 2D data for supervision, and capable of producing high-quality 3D results efficiently~\cite{dong2023ag3d}. Unfortunately, \textit{conditioning} GANs, e.g., on text is far from straightforward since they can be unstable during training, and lack the convenient formalism of score-based (diffusion) generative models, which enable conditioning via classifier or classifier-free guidance \cite{ho2021classifierfree}.

We present \methodname as the first method that successfully trains a feed-forward text-to-3D diffusion-generator using only single-view 2D data for supervision. 
We achieve this by combining the power of unconditional GAN-based generation to enable learning from only 2D data with the power of diffusion models to enable text conditioning. Specifically, we first utilize a GAN to learn a category-specific latent space in the form of triplane features~\cite{chan2022eg3d}, along with the corresponding triplane decoder. This stage requires only single-view 2D supervision. We then distill the extracted GAN latent space into a \textit{text-conditioned} diffusion model by using the unconditional GAN to generate triplane samples, generating corresponding text captions, and training a text-to-triplane diffusion model on this dataset.
%and the corresponding 3D models. We then generate corresponding text captions and train a text-to-triplane diffusion model directly on the sampled triplanes.
Crucially, our method is \textit{feed-forward}, in that it completely avoids test-time optimization, e.g., with Score Distillation Sampling, and instead generates results via learned denoising diffusion in the latent (triplane) space. Overall, \methodname combines the strengths of GANs for efficient training and sample generation with the learned diffusion model, enabling principled and effective conditioning.

We evaluate \methodname in the context of generating 3D human models. We leverage the power of text-to-image foundational models~\cite{Rombach_2022_CVPR} to create a diverse and large-scale synthetic 2D dataset to demonstrate our method. Specifically, we render pose and depth condition images from a deformable human template model~\cite{SMPL:2015} and map these  to realistic and high-quality images conditioned on procedurally generated text prompts via a pre-trained text-to-image model. This enables to create a high-quality dataset that is much larger compared to those commonly used in prior work~\cite{EVA3D,dong2023ag3d}. We compare our approach to possible baselines that can operate with only 2D data for training but leave out methods requiring access to 3D object meshes. %With the exception of the parallel effort~\cite{lei2023diffusiongan3d},
We also evaluate our method for generating real-world faces and cats using the FFHQ and AFHQ dataset~\cite{chan2022eg3d}, as well as generating realistic 3D people using the DeepFashion dataset~\cite{deepfashion}, and demonstrate its compatibility with two GAN architectures, AG3D~\cite{dong2023ag3d} and EG3D~\cite{chan2022eg3d}.

We are unaware of any other generative method that supports feed-forward text-conditioned 3D generation and can be trained with only 2D data. \cref{fig:teaser} shows results obtained using our approach compared to the most closely related baselines. Our method is the only one that can generate high quality \textit{text-guided} 3D models, while avoiding test-time optimization or explicit 3D supervision.

In summary, our key contributions are as follows:
\begin{enumerate}[noitemsep,topsep=0pt]
    \item We highlight the complementary strengths of GANs and denoising diffusion models for 3D generation. The former can be trained effectively with 2D supervision, while the latter allows \textit{guided} generative modeling in the space of learned features.
    \item We propose a novel framework that combines the aforementioned complementary strengths to present a text-guided diffusion-based 3D generator, trained using only single view 2D supervision.
    \item We demonstrate that
    %with appropriate training
    the resulting approach produces high-quality output and avoids costly test-time optimization, outperforming strong recent baselines.
\end{enumerate}

% Figure~\ref{fig:teaser} shows ...

%In summary, ... 

%The goal is to have a feed-forward generator that can generate text-conditioned 3D human representation. 

%Current approaches:
%Train a diffusion model (easier to add text conditioning) in a fashion similar to latent diffusion, but this requires to have access to large scale 3D data where one can first optimize the 3D latent (“triplane”) representation (Rodin)

%Train a generative model, e.g., GANs, from 2D data only (e.g., AG3D) but adding text conditioning to GANs is not straightforward since they are not as stable to train

%Train a diffusion model from 2D images directly (similar to RenderDiffusion) but this does not generate high quality since it’s a challenging problem, the diffusion model has to “jointly” learn the latent space and the input (2D image) and the output (triplane) of the diffusion model are not same representation creating a quality bottleneck (edited) 

%Our approach: We combine the power of unconditional GAN based generation to enable learning from only 2D data with the power of diffusion model to enable text conditioning. In other words, we utilize the GAN to first learn the latent space, i.e., “triplanes” together with the corresponding triplane decoder. Then we distill the knowledge of the GAN into a text conditioned diffusion model. We do this by sampling the unconditional GAN to generate many triplanes for which we can generate text captions and then train a diffusion model directly on the triplanes.

\section{Related Work}
\paragraph{3D generation with 3D data}
Following the success of 2D generation models in high quality image generation~\cite{Karras2019stylegan2,Rombach_2022_CVPR}, there have been various attempts to learn 3D generation models. A wide variety of methods have been proposed using 3D data as supervision. Different generator architectures such as GANs~\cite{li2021spgan,Xie_2021_CVPR}, normalizing flows~\cite{pointflow}, and more recently diffusion models~\cite{zeng2022lion,shue20233d} have been extended to the 3D domain. Unlike images that are universally represented as 2D pixel arrays, 3D data can be represented in various forms. Hence, 3D generators have also explored different options such as point clouds~\cite{achlioptas2018learning,ShapeGF}, voxels~\cite{Wu16}, implicit representations~\cite{Park_2019_CVPR}, and radiance fields~\cite{gaudi,hyperdiffusion}. While promising, these methods have been limited because the amount of 3D data is orders of magnitude less than the image counterparts (despite recent datasets such as Objaverse~\cite{objaverse}). 

\vspace{-0.8em}
\paragraph{3D generation with 2D data}
To tackle the 3D data scarcity problem, a parallel line of work focuses on leveraging 2D data for 3D generation. In this context, 3D-aware GANs~\cite{chanmonteiro2020pi-GAN,deng2022gram} have been in particular effective where a GAN is used to generate an implicit representation~\cite{henzler2019platonicgan,Niemeyer2020GIRAFFE,Schwarz2020NEURIPS,xiang2023gramhd,gao2022get3d} including a NeRF~\cite{gu2022stylenerf}, a volumetric field~\cite{Nguyen-Phuoc2019a}, or a triplane~\cite{chan2022eg3d}. These representations are rendered from different viewpoints for which an image-based discriminator is used for adversarial supervision. Our method leverages the power of such unconditional generators in the first stage. However, we demonstrate that it is not trivial to add text conditioning to such methods and hence we resort to diffusion models in the second stage. %Get3D~\cite{gao2022get3d} further extracts a mesh from the implicit representation, which is then rendered with a texture. 
More recently, there have been efforts to train 3D diffusion models using 2D images only. HoloDiffusion~\cite{karnewar2023holodiffusion} creates feature grids from input videos which are then used to generate different multi-view images via a diffusion model. The follow up work, HoloFusion~\cite{karnewar2023holofusion}, extends this setup with a superresolution module. RenderDiffusion~\cite{anciukevivcius2023renderdiffusion} generates a triplane representation from a single image using reconstruction losses. While promising, such diffusion based 3D generators trained on 2D data only have not reached the quality of their GAN counterparts. Hence, our method leverages a 3D-aware GAN to generate 3D samples for training a text-conditioned diffusion based generator.

\vspace{-1.2em}
\paragraph{Text-to-3D generation}
Breakthrough advances such as CLIP~\cite{RadfordKHRGASAM21} have proposed ways to obtain common latent spaces between text and image modalities, enabling text-guided generation and editing applications. Several works~\cite{Gao_2023_SIGGRAPH} have utilized loss functions in the CLIP embedding space directly for text-guided 3D generation and editing. The seminal work of DreamFusion~\cite{poole2022dreamfusion} distills a text-to-image model to generate 3D objects using \emph{Score Distillation Sampling} (SDS), with follow-ups improving 
%has demonstrated that the knowledge embedded in a large scale text-to-image generation model can also be distilled in an optimization using \emph{Score Distillation Sampling} (SDS) to generate 3D objects from text prompts.
%Several follow up works have improved on this framework both in terms of 
quality~\cite{lin2023magic3d,chen2023fantasia3d} and optimization efficiency~\cite{lorraine2023_att3d}. Other work~\cite{gu2023learning} also leverages a GAN to obtain a latent triplane representation, but requires test-time optimization for text conditioning. Despite the advances, these approaches still require a per-prompt optimization step, which can be costly thus limiting their efficiency and scalability. 
Another line of work~\cite{hong2023lrm, liu2023syncdreamer, liu2023one2345} generate multi-view images that can be reconstructed into 3D objects without optimization, by fine-tuning text-to-image models. However, fine-tuning requires synthetic 3D data that introduces bias and reduces the generality of these methods.

% \vspace{-1.1em}
\paragraph{Human-specific 3D generation}
In the context of humans, statistical template models such as SMPL~\cite{SMPL:2015} can be considered as early examples of 3D generators, which are learned from collections of human scans with minimal/tight clothing. For with clothing, Cape~\cite{ma2020cape} trains a graph CNN architecture from scans of clothed humans. With the increasing effectiveness of implicit representations, follow up work has presented neural implicit generators~\cite{chen2022gdna,palafox2021spams,Palafox2021NPMsNP}. More recently, in an impressive effort, Rodin~\cite{Wang2022RODINAG} has presented a diffusion-based architecture to generate text-conditioned humans. 
The above models, however, require access to 3D human data, and unfortunately, existing datasets~\cite{2023dnarendering,cai2022humman} are limited in quantity and diversity. Hence, methods like Get3DHuman~\cite{xiong2023Get3DHuman} propose using pseudo ground truth labels obtained from single image human reconstruction methods to overcome the data challenge. In another line of work, researchers~\cite{sun2022ide,dong2023ag3d} have trained 3D aware GANs specifically on human image datasets. While showing impressive results, these methods are not straightforward to extend to accept new guidance or conditioning, e.g., via text prompts, due to the known stability issues in GAN training. Hence, our method takes a different route and utilizes the knowledge captured in such a GAN framework, i.e., AG3D~\cite{dong2023ag3d}, to train a text-conditioned diffusion model. 
Finally, there has been a series of efforts~\cite{jiang2023avatarcraft,huang2024tech} that extend the SDS based optimization approaches to avatar generation. Such methods utilize a deformable template body such as SMPL~\cite{kim2023chupa} or GHUM~\cite{kolotouros2023dreamhuman} to regularize the optimization process. More recently, approaches that also exploit additional conditioning strategies for the base image generation model, such as ControlNet~\cite{zhang2023avatarverse,cao2023dreamavatar}, have been proposed. Our method also takes advantage of a ControlNet architecture to generate a large set of human images. However, we use these images to train a \textit{feed forward} generation model instead.
In a concurrent effort, DiffusionGAN3D~\cite{lei2023diffusiongan3d} performs domain adaption of a 3D aware GAN (e.g., EG3D~\cite{chan2022eg3d}) using SDS loss and enables text conditioning by searching for a latent code in the latent space at test time. In contrast, our method directly uses the GAN latent space to train a text-guided diffusion model and avoids any test time optimization.

% \vspace{-1em}
\paragraph{3D generation in a learned latent space}
Finally, we note that our approach is related to existing methods that first pre-train a latent representation for 3D data and then train diffusion models in this learned space, e.g., \cite{zeng2022lion,hyperdiffusion,shue20233d,ntavelis2023autodecoding,nam20223d,gupta20233dgen,chou2023diffusion}. However, these approaches
%(with the exception of \cite{gu2023learning}, which is \textit{concurrent} to our work)
are based on an \textit{auto-encoding strategy} and build the latent space using explicit 3D or multi-view supervision. Instead, we only use unorganized \textit{single view 2D data} without camera poses for supervision. This is a significantly harder problem as it lacks explicit 3D supervision, but at the same time allows us to greatly enlarge the training corpus. Overall we demonstrate that our approach leads to an efficient feed-forward text-guided 3D generative model that does not require \textit{any} 3D training data.

\begin{figure*}[t]
    \includegraphics[width=\textwidth]{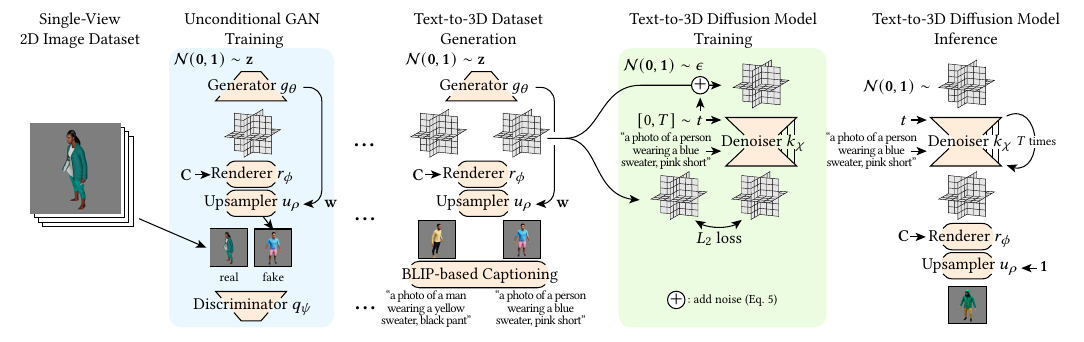}
    \caption{\textbf{Method Overview.} We train a feed-forward text-to-3D diffusion model in two stages. First, we train an unconditional GAN-based 3D object generator like AG3D~\cite{dong2023ag3d} with a single-view image dataset. 3D objects are represented as triplanes and rendered with a renderer followed by an upsampler. We generate a large set of triplanes and caption them using BLIP~\cite{li2022blip}. The resulting (triplane, caption) dataset is then used to train a text-to-3D diffusion model, effectively distilling the GAN generator into a diffusion model, while also allowing for text conditioning. Triplanes generated by the text-to-3D model are rendered using the renderer and upsampler trained in the first stage.}
    \label{fig:overview}
    \vspace{-1.2 em}
\end{figure*}

\section{Motivation and Overview}
\label{sec:motivation}
Our approach is motivated by several considerations. First, as mentioned above, our key objective is to design a method capable of generating text-conditioned 3D geometry, while using only 2D images for supervision. Furthermore, we aim to avoid costly test-time optimization, e.g., via Score Distillation Sampling, and instead perform synthesis through standard denoising diffusion at inference. A key question when applying denoising diffusion is deciding in \textit{which space} diffusion is performed. This question is especially prominent in a setup like ours since our supervisory signal (2D images) differs from the target output (3D geometry), while denoising diffusion is typically performed in some fixed space.

One general possibility is to train an image-based encoder to produce latent features, from which 3D geometry can be generated while performing denoising diffusion in this latent space. Perhaps the most natural option would be to train the entire network end-to-end, with both the feature encoder-decoder as well as the latent denoising networks using, e.g., a rendering-based loss. This option was recently explored in RenderDiffusion~\cite{anciukevivcius2023renderdiffusion} where a denoising network learns to simultaneously denoise input images, encode triplane features that are used
%and denoise in the space of learned triplane features is used
to generate 3D geometry, and decode triplane features for rendering. Unfortunately, as we demonstrate in \cref{sec:results}, we observe that such an approach has limited accuracy, which we attribute to three key reasons: (i)~training \textit{both} denoising diffusion \textit{and} triplane encoding tends to be error-prone, given the very significant degrees of freedom; (ii)~due to the mismatch in the input (images) and output (triplane features), the UNet architecture cannot exploit skip-connections, forcing the denoising network to operate purely on the bottleneck representation, thus significantly limiting the quality of the output; and finally, (iii)~input and rendered viewpoints need to match, making it possible for this approach to overfit to the input viewpoints and produce triplanes that are lower quality when rendered from other viewpoints.

To overcome these challenges, we combine the strengths of two major approaches. First, following \cite{dong2023ag3d} we use a GAN to train a latent triplane-based representation using single view 2D images as supervision, but from which 3D geometry can be generated. Importantly, training a GAN does not require the input and output to be in the same space. Moreover, it also does not assume access to rendering viewpoints, which it could potentially overfit to, making it well-adapted for our scenario.

% \maks{Maybe here the possibility of cheating? }

On the other hand, \textit{conditioning} GANs with text is known to be challenging, owing in part to the lack of interpretation via score matching enjoyed by diffusion models, which enables, e.g., classifier-free guidance \cite{ho2021classifierfree}. Thus, after training the GAN as a triplane generator, we train a separate diffusion model operating on a set of generated triplanes.

This two-stage training with a GAN and a diffusion model allows text conditioning and leads to high-quality results.
In the following sections, we provide the details of our method and compare it to strong recent baselines.

% \vspace{-0.6em}
\section{Method}
% \subsection{Method Overview}
%
As mentioned above, our goal is to train a feed-forward generator for 3D objects using single-view 2D images $\mathcal{I} = \{I_1, \dots, I_N\}$ as supervision, with $I \in \mathbb{R}^{3 \times H \times W}$.
We define 3D objects as neural fields that encode densities and colors, and use triplanes~\cite{peng2020convolutional,chan2022eg3d} as representation for these fields. This representation is described in \cref{sec:scene_representation}.

To reach our objective, we proceed in two training stages (see \cref{fig:overview}):
%
%We represent scenes using triplanes~\cite{peng2020convolutional, chan2022eg3d} and proceed in two stages to train our feed-forward text-to-3D generator
%We proceed in two training stages (see Figure~\ref{fig:overview}):
%
First, we train a GAN as an \textit{unconditional} generator for triplane representations of 3D objects, using the 2D image dataset $\mathcal{I}$ as supervision. Previous work has shown that class-specific GANs can be trained successfully to generate 3D objects with only 2D supervision~\cite{chan2022eg3d, dong2023ag3d}. We use the architecture proposed in AG3D ~\cite{dong2023ag3d} for this step. This unconditional generator produces high-quality 3D objects, but we found it hard to introduce text conditioning into a GAN architecture, due to the inherently unstable training process. This observation is supported by the absence of text-conditioned GANs in recent literature, with the exception of a few carefully tuned 2D GANs~\cite{zhou2022lafite, Sauer2023stylegant, kang2023gigagan}. Using the trained GAN, we then create a large dataset of 3D objects and caption them using BLIP~\cite{li2022blip}. See \cref{sec:gan} for details on the first stage.

In the second stage, we train a text-conditioned diffusion model on the dataset generated in the first stage, using an architecture based on StableDiffusion~\cite{Rombach_2022_CVPR}. We remove the encoder, as our 3D objects have already been encoded into triplanes by the GAN, and carefully normalize the triplanes. We found that training the diffusion model with 3D objects encoded via triplane features, as supervision, rather than using 2D supervision as in RenderDiffusion~\cite{anciukevivcius2023renderdiffusion}, is essential to obtain high-quality outputs. We show a comparison to a RenderDiffusion-based setup in \cref{sec:results}. We describe the details of the second stage in \cref{sec:diffusion}.

% We observe that neither of the existing generative approaches alone are particularly suitable for this task. GANs are amenable to training with 2D supervision 

\subsection{3D Object Representation}
\label{sec:scene_representation}
We define 3D objects as neural fields $s$, consisting of an RGB color field representing the albedo, and a pseudo Signed Distance Function (SDF) representing the geometry of the 3D object. The neural field can be queried at a position $x \in \mathbb{R}^3$ to give a pseudo-SDF value $d$ and an albedo color $\mathbf{c}$ at that position: $(d, \mathbf{c}) = s(x)$. 

We use a triplane representation~\cite{peng2020convolutional, chan2022eg3d}, consisting of 3 orthogonal 2D feature grids $\mathbf{T}_{xy}, \mathbf{T}_{xz}, \mathbf{T}_{yz}$, one for each coordinate plane, with $\mathbf{T}_{**} \in \mathbb{R}^{n \times h \times w}$. The three feature grids can conveniently be concatenated into a single multi-channel image $\mathbf{T} \in \mathbb{R}^{3n \times h \times w}$. A feature $\mathbf{T}(x) \in \mathbb{R}^{3n}$ for a given query point $x$ is obtained by projecting $x$ to each coordinate plane, querying the corresponding feature grid with bilinear interpolation, and concatenating the three resulting feature vectors. This feature is converted to a pseudo-SDF and color value using an MLP $h_\phi$ with parameters $\phi$ as decoder:
\begin{equation}
    (d, \mathbf{c}) = s(x) \coloneqq h_\phi(\mathbf{T}(x)).
\end{equation}

\vspace{-0.2em}
\textbf{3D object rendering}
We use a volumetric renderer based on previous work~\cite{mildenhall2021nerf, chan2022eg3d, dong2023ag3d}. Samples are placed along each camera ray using a two-pass strategy: stratified sampling followed by importance sampling. Samples along a ray $\mathbf{v}$ are then accumulated as:
\begin{align}
    \label{eq:ray_accumulation}
    \mathbf{c}^{\mathbf{v}} =& \sum_i \mathbf{c}_i^{\mathbf{v}} a_i^{\mathbf{v}} \prod_{j<i}(1-a_j^{\mathbf{v}}),\\
    \text{with }a_i^{\mathbf{v}} =&\ 1 - \exp(-\sigma_i^{\mathbf{v}} \delta_i^{\mathbf{v}}), \nonumber \\
    \sigma^\mathbf{v}_i =&\ \text{sigmoid}(d^\mathbf{v}_i),\nonumber \\
    (d^\mathbf{v}_i, \mathbf{c}^\mathbf{v}_i) =&\ h_\phi(\mathbf{T}(x^\mathbf{v}_i)). \nonumber 
\end{align}
%
% r_\phi(\mathcal{V},\mathbf{T})
The sample locations $x^\mathbf{v}_i$ along ray $\mathbf{v}$ are indexed from the camera outwards. At each sample location, $\sigma^{\mathbf{v}}_i$ are computed from the pseudo-SDF values $d^{\mathbf{v}}_i$ using a sigmoid function, and $\delta_i^{\mathbf{r}}$ is the local spacing between samples along the ray. For convenience, we also define a rendering function $r$ that outputs a rendered RGB image given a triplane $\mathbf{T}$ and camera parameters $\mathbf{C}$:
\begin{equation}
    \label{eq:renderer}
    r_\phi(\mathbf{T}, \mathbf{C}) = [ \mathbf{c}^\mathbf{v} ]_{\mathbf{v} \in \mathcal{V}_\mathbf{C}},
\end{equation}
where $\mathcal{V}_\mathbf{C}$ is the set of rays for camera $\mathbf{C}$ and the square brackets denote concatenation and reshaping into an image $r_\phi(\mathbf{T}, \mathbf{C}) \in \mathbb{R}^{3 \times H \times W}$. Note that the rendering function depends on the learned parameters $\phi$ of the decoder $h_\phi$.

\subsection{Training an Unconditional GAN for 3D Objects}
\label{sec:gan}
Given a latent vector sampled from the standard normal distribution $\mathbf{z} \sim \mathcal{N}(\mathbf{0},\mathbf{1})$, the unconditional GAN $g_\theta$ creates a triplane representation of a 3D object: $\mathbf{T} = g_\theta(\mathbf{z})$.

\paragraph{Architecture}
We use an architecture based on AG3D~\cite{dong2023ag3d}. It consists of a generator $g_\theta$, a renderer $r_\phi$, an upsampler $u_\rho$, and a discriminator $q_\psi$. The generator and discriminator are based on StyleGAN2~\cite{Karras2019stylegan2} and the upsampler on EG3D~\cite{chan2022eg3d}.

\vspace{-0.8em}
\paragraph{Rendering}
To render a triplane $\mathbf{T}$, we use the renderer $r_\phi(\mathbf{T}, \mathbf{C})$. As high-resolution renders have prohibitive time and memory requirements, we render in low resolution ($128^2$) and use the upsampler $u_\rho$ to get higher-resolution renders ($256^2$). The upsampler uses two inputs in addition to a low-resolution color render: i) a low-resolution render of triplane features, providing additional information about local details of the 3D object, and ii) a style vector $\mathbf{w}$ from the StyleGAN-based generator, providing global information about the entire 3D object:
\begin{equation}
    I^\text{hi} = u_\rho\big(r_\phi(\mathbf{T}, \mathbf{C}),\ r(\mathbf{T}, \mathbf{C}),\ \mathbf{w}\big),
\end{equation}
where $I^\text{hi}$ is the high-resolution image. Triplane features $r(\mathbf{T}, \mathbf{C})$ are rendered by replacing the color values $\mathbf{c}$ in \cref{eq:ray_accumulation} and~\cref{eq:renderer} with sampled triplane features $\mathbf{T}(x)$. The style $\mathbf{w}$ is an intermediate result of the generator $g_\theta$, a non-linear transformation of the latent vector $\mathbf{z}$.

% Images are rendered at $256^2$ and upsampled to $512^2$.

\vspace{-0.8em}
\paragraph{Training}
The generator, renderer, upsampler, and discriminator are trained jointly with the standard adversarial loss described in StyleGAN2~\cite{Karras2019stylegan2}, using single-view 2D training from $\mathcal{I}$ as real images and upsampled renders $I^\text{hi}$ from the generated objects as fake images.
Random viewpoints are used for the camera parameters $\mathbf{C}$.
%
% Additionally, \paul{TODO: additional losses from AG3D}

As demonstrated, e.g., in \cite{henzler2019platonicgan,dong2023ag3d} GANs can be trained as 3D generators without 3D or multi-view supervision. The adversarial loss does not require ground truth for any specific render, and instead only requires the distribution of renders in $\mathcal{I}^R$ to match the distribution of images in $\mathcal{I}$.
%does not require explicit ground truth for any of the renders. Instead, we only require that the distribution of renders $\mathcal{I}^R$ matches the distribution of images in $\mathcal{I}$.
This allows generating objects unconditionally at training time and rendering them from arbitrary viewpoints, even if we do not have ground truth for these specific objects or viewpoints.

% \subsection{Training a Text-to-3D Diffusion Model for 3D Objects}
\subsection{Text-to-3D Diffusion Model for 3D Objects}
We use the GAN generator $g_\theta$ to create a dataset of triplanes $T$ and caption them using BLIP~\cite{li2022blip}. Details are described in ~\cref{sec:results}. We then use this dataset of (caption, triplane) pairs to train a text-conditioned diffusion model as our feed-forward text-to-3D generator, effectively distilling the unconditional distribution learned by the GAN into a text-conditioned version of the distribution.

\textbf{Triplane pre-processing}
Before training on our dataset of triplanes, we normalize the triplanes by subtracting a per-channel mean and dividing by a per-channel standard deviation. Both mean and standard deviation are computed over the full dataset of triplanes. Additionally, we remove outlier values by scaling triplanes with a factor of $1/16$ and clamping to $[-1,1]$.

\textbf{Training and Architecture}
For each training sample, we randomly pick a triplane $\mathbf{T}^0$ and a corresponding caption $Y$ from the dataset. We then add noise to a triplane according to a noise schedule $\alpha_t$:
\begin{equation}
    \label{eq:add_noise}
    \mathbf{T}^t = \sqrt{\alpha_t}\ \mathbf{T}^0 + \sqrt{1-\alpha_t}\ \epsilon \text{ with } \epsilon \sim \mathcal{N}(\mathbf{0}, \mathbf{1}),
\end{equation}
The noise schedule $\alpha_t$ determines the amount of noise added to the triplane given a time step $t$. We use a Sigmoid noise schedule~\cite{allan2023sigmoidsched}, which has shown good performance on high-resolution images. For each training sample, we randomly pick the time step $t$ with uniform probability in $[0,T]$. Given the noisy image, we train a denoiser network $k_\chi$ with parameters $\chi$ to predict the denoised image using an $L_2$ loss:
\begin{equation}
\mathcal{L}_{\text{diff}} = \|k_\chi(\mathbf{T}^t, Y, t) - \mathbf{T}^0\|^2_2.
\end{equation}
In addition to the noisy triplane, the denoiser takes as input the text prompt $Y$ and the current time step $t$.
We randomly replace $Y$ with empty text with $20\%$ probability to allow for unconditional generation when $Y$ is empty; this is required to support classifier-free guidance at inference time.
The denoiser $k_\chi$ is implemented as a UNet with the same architecture as in StableDiffusion~\cite{Rombach_2022_CVPR}.

\vspace{-0.8em}
\paragraph{Inference}
Given a triplane sampled from a pure noise distribution $\mathbf{T}^T \sim \mathcal{N}(\mathbf{0},\mathbf{1})$ and a text caption $Y$, we predict a denoised triplane $\mathbf{T}^0$ with multiple denoising iterations using the denoiser $k_\chi$. In each iteration, the denoiser outputs a prediction of the denoised image $\hat{\mathbf{T}}^t \coloneqq k_\chi(\mathbf{T}^t, Y, t)$. We use this prediction in a deterministic DDIM sampler~\cite{song2021ddim}:
\begin{equation}
\mathbf{T}^{t-1} = \sqrt{\alpha_{t-1}}\ \hat{\mathbf{T}}^t + \sqrt{1- \alpha_{t-1}}\ \hat{\epsilon}^t,
\end{equation}
where $\hat{\epsilon}^t$ is the predicted noise:
\begin{equation}
\hat{\epsilon}^t \coloneqq (\mathbf{T}^t - \sqrt{\alpha_t}\ \hat{\mathbf{T}}^t)\ /\ \sqrt{1-\alpha_t},
\end{equation}
and $\alpha_t$ is the noise schedule.
%that determines the amount of noise added to the triplane in each step. We a use a Sigmoid noise schedule~\cite{allan2023sigmoidsched}, which has shown good performance on high-resolution images.
%
% k_\chi(\mathbf{T}^t, Y, t) + \sqrt{1 - \alpha_{t-1}} \frac{\mathbf{T}^t - \sqrt{\alpha_t} k_\chi(\mathbf{T}^t, Y, t)}{\sqrt{1-\alpha_t}}
We additionally use classifier-free guidance~\cite{ho2021classifierfree} with a guidance scale of $7.5$ to better align generated triplanes with text prompts.
% \paul{TODO: update classifier-free guidance description once experiments are finished.}

\vspace{-0.8em}
\paragraph{Rendering}
At inference time, we render triplanes using the renderer $r_\phi$ and upsampler $u_\rho$ that was trained by the GAN. While the upsampler $u_\rho$ requires the style vector $\mathbf{w}$ as input, which is not available in this setting, we found that the upsampler is not sensitive to this input. We pass a vector of all-ones, giving us similar upsampling quality.

\label{sec:diffusion}

\section{Results}
\label{sec:results}

% \subsection{Data Processing}
\textbf{Single-view image dataset.} 
To obtain a sufficiently large training set of 2D images that we can use both for our method and for baselines, we create the dataset of synthetic images with StableDiffusion 1.5, conditioned on depth maps, pose maps, and prompts. To obtain depth and pose maps, we take the 3D human body model SMPL~\cite{SMPL:2015} in A-pose and render a depth map as well as a 2D pose map from a viewpoint sampled around the 3D body. In practice, we use a fixed distance of $2.34$ to the origin, a fixed elevation of $40$ degrees, and randomly sample an azimuth.
%sample viewpoints randomly viewpoints around the body (add details about azimuth, elevation).
%We use the rendered depth and pose maps as conditioning signals to corresponding ControlNets~\cite{zhang2023adding} applied on Stable Diffusion 1.5. 
To obtain the prompts, 
%Although not necessary for the first stage of our method, some baselines that we compare against require (text, image) pairs for training. For this,
we use a procedural approach that is aimed to describe the appearance of a person. In particular, we generate prompts that include a gender keyword, types of upper and lower-body clothing, and their corresponding colors. We select these clothing types and colors from a predefined set, which consists of 28 upper-body clothing types, 12 lower-body clothing types, 11 footwear types, and 23 colors. Given such a prompt and the depth/pose map renderings, we generate the corresponding RGB image resulting in a dataset of 300k images in total. A few examples are shown in \cref{fig:dataset}. %the supplementary materials. %Figure~\ref{fig:dataset}.
% \begin{figure*}[t]
%     \includegraphics[width=\textwidth]{images/dataset_examples.pdf}
%     \caption{\textbf{Dataset examples.} We show a few images and corresponding prompts from our single-view image dataset. Prompts that were initially used to generate the images are often not accurate, we refine them with our BLIP-based captioning approach.}
%     \label{fig:dataset}
%     % \vspace{-3em}
% \end{figure*}
%
\begin{figure*}[t]
    \includegraphics[width=\textwidth]{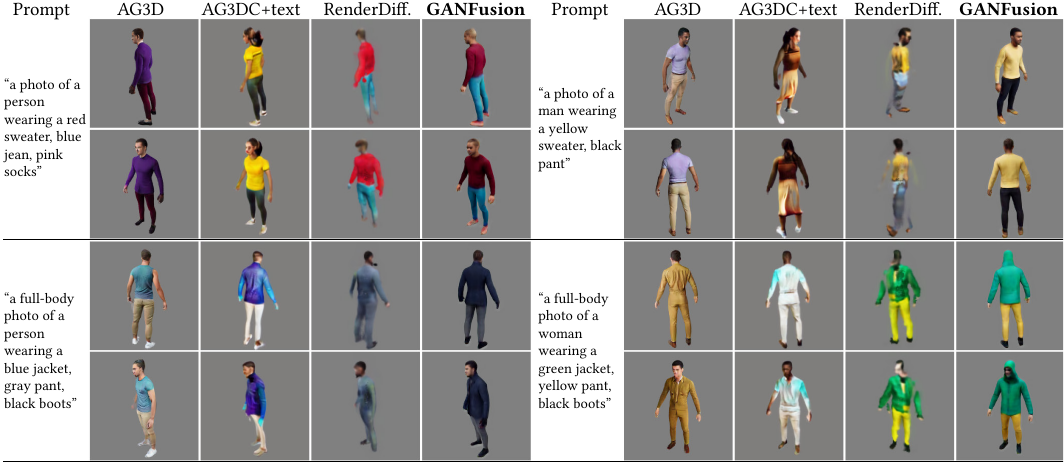}
    \caption{\textbf{Qualitative comparison.} We provide visual results from our method as well as the baselines. We provide random generations from AG3D~\cite{dong2023ag3d} which is unconditional. While random samples are of high quality, it is not straightforward to extend AG3D to enable text conditioning even with the addition of CLIP based losses during training. RenderDiffusion~\cite{anciukevivcius2023renderdiffusion} can follow the text prompts but lacks in terms of quality. \methodname matches the quality of AG3D while enabling text conditioning. For AG3D and our results, we visualize the results with the 2D upsampler trained jointly with the 3D GAN generator in the first stage of our method.}
    \label{fig:comparison}
    \vspace{-1.4em}
\end{figure*}
While Stable Diffusion generates a plausible image, we observe that it cannot perfectly follow the text prompt. For example, the generated image might depict a different clothing type or color than instructed in the prompt. While we do not need text prompts matching the images for our method, the baselines we compare to do require matching prompts. Hence, in a second pass, we refine the captions of the generated images using the same BLIP-based~\cite{li2022blip} captioning we use to caption the generated triplanes (described below). We emphasize that these prompts are \textit{not} used in the first stage of training of our method.
Finally, we also generate an additional set of $1k$ (prompt, image) pairs that we use for evaluations using the described procedural technique.

\textbf{Triplane dataset.}
To train the diffusion model in the second stage of our method, we use the GAN generator $g_\theta$ trained in stage 1 to create $49.5k$ triplanes. We render these triplanes from random viewpoints following the same viewpoint distribution as used in the 2D dataset and caption them using BLIP-based captioning.

\textbf{BLIP-based captioning.}
We create captions from images of persons using a visual question-answering approach based on BLIP~\cite{li2022blip}. We define a set of questions regarding the gender and clothing of a person (for example, \emph{``What is the person wearing on their feet?''}) and also define a set of possible answers for each question for BLIP to choose from (for example, \emph{``sneakers''}, \emph{''loafers''}, \emph{``boots''}, etc.). BLIP returns the best fitting answers for each of the questions, which we use to construct the prompt by filling in blanks in a template. \cref{sec:blip_labeling} provides additional details.

% use the answers to construct a prompt
% \paul{TODO: describe (can be moved to supp. if we need space)}
% \subsection{Comparisons}
\textbf{Baselines.}
We compare our method to (i)~\emph{AG3D}~\cite{dong2023ag3d}, which we adopt in the first part of our method. (ii)~We attempt to extend AG3D directly with additional text conditioning (\emph{AG3D+text}), using an approach based on cross-attention layers following the implementation of GigaGAN~\cite{kang2023gigagan}. However, due to GAN training instabilities, we were unable to get a working version of this baseline. To improve stability, we introduce (iii)~a modified version of AG3D+text, called \emph{AG3DC+text} where we encourage adherence to the text prompt by adding a CLIP loss between the text prompt and the rendered generator output instead of conditioning the discriminator with the prompt. (iv)~We compare to \emph{RenderDiffusion}~\cite{anciukevivcius2023renderdiffusion} where we train a diffusion network to directly predict the triplane parameters. While training this baseline, given an image from our dataset and the corresponding text prompt, we add noise to it based on a sampled diffusion timestep $t$ and predict the final triplane features directly conditioned on the prompt. We then render the triplane from the corresponding viewpoint and define an L1-based reconstruction loss. In this baseline, the triplane decoder is jointly trained with the diffusion model. We train all the baselines on the same dataset as our method. 
To perform a more fair evaluation of the triplane generation quality, we show results both with and without the image upsampler component of AG3D and our method.
%Later in the section, we demonstrate how our results can also be upsampled using a similar superresolution technique.
Note that AG3D used a more realistic dataset, as reflected in their results, while we opt for a synthetic dataset to obtain more diverse prompts that better demonstrate text conditioning.
% It should be noted that, unlike the dataset used in the AG3D paper which features more real-world images, our dataset is primarily synthetic. This design choice ensures a diverse range of text prompts to better demonstrate text conditioning while maintaining a manageable size.}

\textbf{Metrics.}
We use two quantitative metrics. First, the Frechet Inception Distance (FID) to measure the fidelity of the generation results. Specifically, treating our generated image dataset as the real distribution, we use InceptionV3~\cite{szegedy2016rethinking} to obtain features. 
Second, we use CLIP similarity to measure the text alignment between the input prompts and generations, computed as the cosine similarity between the CLIP embeddings of text prompt and rendered outputs. For both metrics, we render images from random viewpoints, following the same viewpoint distribution used in the 2D dataset.

% Specifically, we render the generated triplanes (add details about the viewpoint) and compute the similarity between the input prompts and the rendered images in the CLIP space.

\begin{figure*}[t]
    \includegraphics[width=\textwidth]{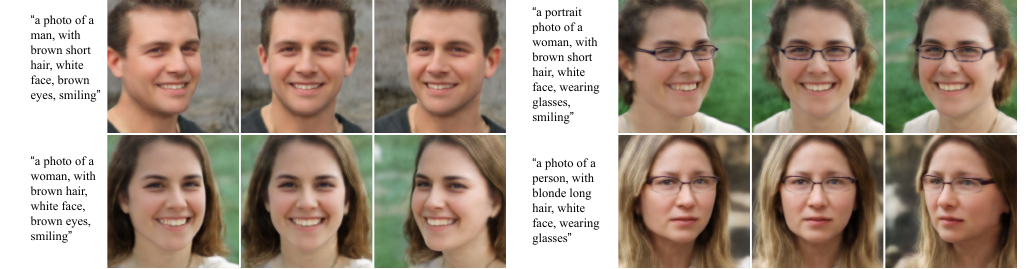}
    \caption{\textbf{Qualitative results on the FFHQ dataset~\cite{chan2022eg3d}.} We replace AG3D~\cite{dong2023ag3d} with EG3D~\cite{chan2022eg3d} as the generator in our first stage to effectively enable text-conditioning on real-world face data.}
    \label{fig:ffhq}
    % to modify:
    % https://docs.google.com/presentation/d/18flXl7elkANLNBva5jrEFem-6ts60SkV5vrco8oBJdw/edit?usp=sharing
    \vspace{-0.9em}
\end{figure*}

% \vspace{-0.5em}
% \subsection{Evaluation}

\begin{table*}[t]
    \scriptsize
    \centering
    \ra{1.0}
    
    % \vspace*{.05in}
    \resizebox{\textwidth}{!}{
    \begin{tabular}{@{}rccccccc@{}}  %% {\textwidth}
         \renewcommand{\arraystretch}{0.5}
         % \toprule
         % & AG3D (low) & AG3D (high) & AG3D+text & RenderDiff. & \textbf{\methodname} (low) & \textbf{\methodname} (high) \\
         & \multicolumn{2}{c}{Unconditional} & \multicolumn{5}{c}{Text Conditioned} \\
         \cmidrule(lr){2-3} \cmidrule(lr){4-8}
         & AG3D (low) & AG3D (high) & AG3D+text & AG3DC+text & RenderDiff. & \textbf{\methodname} (low) & \textbf{\methodname} (high) \\
         \midrule
         FID$\downarrow$ &  \multicolumn{1}{|c}{104.3} & \multicolumn{1}{c|}{\textbf{35.8}} & 289.6 & 92.0 & 135.7 & 88.23 & \textbf{68.8} \\
         CLIP sim.$\uparrow$ & \multicolumn{1}{|c}{$\times$} & \multicolumn{1}{c|}{$\times$} & 0.174 & 0.240 & 0.263 & \textbf{0.296} & \textbf{0.293} \\
         % \bottomrule
    \end{tabular}
    }
    \caption{\textbf{Quantitative comparisons.} We compare our method to the unconditional generator AG3D~\cite{dong2023ag3d}, a version of AG3D where text conditioning is added, and the RenderDiffusion~\cite{anciukevivcius2023renderdiffusion} baseline. For the AG3D and our method, we evaluate the generation quality both with (high) and without (low) the use of the upsampler. Our method achieves good generation quality while also enabling text conditioning.}
    \label{tab:comparison}
    \vspace{-1.5 em}
\end{table*}

\textbf{Qualitative comparisons.}
We show a qualitative comparison to all baselines in \cref{fig:comparison} (except the unsuccessful text-conditioned version of AG3D, which we show in \cref{fig:conditional_ag3d}). We can see that AG3D produces high-quality results, but being an unconditional generator, it cannot adhere to the input prompts. The CLIP loss in AG3DC introduces minimal prompt adherence but decreases quality slightly. RenderDiffusion better follows the prompt but suffers from low quality and view inconsistency resulting in artifacts and missing parts when viewed from different directions. As discussed in~\cref{sec:motivation}, we attribute the lack of quality to the challenges RenderDiffusion faces when training a single denoiser network to jointly learn the tasks of (i)~triplane encoding from a 2D image, and (ii)~denoising the 2D image. Our method follows the prompt as closely as RenderDiffusion, but with significantly improved quality, getting close to the quality of the unconditional AG3D.
As RenderDiffusion does not use an image upsampler, we also compare the three methods without upsampling, which are provided in \cref{fig:comparison_without_upsampling}. %the supplementary materials.% in Figure~\ref{fig:comparison_without_upsampling}. 
While this reduces the image detail in both AG3D and our method, both still perform far better than RenderDiffusion. Also, it can be seen that our distillation actually improves the FID compared to AG3D without the upsampler, while, at the same time, enabling text conditioning. We attribute this improvement to the fact our distillation helps to remove low-quality outliers from the distribution. In contrast, since we do not retrain the upsampler it has less benefit after our distillation.
In \cref{sec:add_results}, we extend the qualitative comparisons, and include the text-conditioned AG3D.
%, which was excluded from the previous figures due to all experiments with it resulting in mode collapse or other forms of training instability.
% and provide a more details on the BLIP-based captioning process we used.

% Results for the text-conditioned AG3D are shown in Figure~\ref{fig:conditional_ag3d}. Due to the training instability, all experiments with text-conditioning resulted in mode collapse or other forms of training instability. In the figure, we see a case of severe mode collapse.

\textbf{Results on FFHQ.} We also tested our method on a real-world human face dataset, FFHQ~\cite{chan2022eg3d}. For the unconditional GAN training, we adopted the EG3D architecture~\cite{chan2022eg3d} instead of AG3D. Qualitative results are provided in \cref{fig:ffhq}. Our method successfully enables text conditioning, with a quality similar to the original EG3D paper.
This substituion of AG3D with EG3D demonstrates the versatility of our method to use different 3D GAN architectures that are trained with single-view data, as it is agnostic to the methods used in the initial training stage. Additionally, this suggests that our method can handle both 
%It is evident that our method is capable of handling various types of data, both
synthetic and real-world data. %Additionally, it is agnostic to the methods used in the initial training stage.
In \cref{sec:add_results}, we provide \textbf{additional results} on the AFHQ~\cite{chan2022eg3d} dataset (real-world cat faces), as well as the DeepFashion~\cite{deepfashion} dataset (realistic 3D human figures).

% \vspace{-0.2em}
\textbf{Quantitative comparisons.}
\cref{tab:comparison} provides a quantitative comparison of all baselines on our test set of 1k images. We divide the table into unconditional methods (AG3D with and without upsampling) and text-conditioned methods. The results support our findings from the qualitative results: AG3D with upsampler achieves the best quality (low FID), but is unconditional and has, therefore, no prompt adherence compared to the text-conditioned methods. Naively adding text conditioning to AG3D results in training instability, giving both low quality and prompt adherence. Introducing a CLIP loss in AG3DC lowers the quality and only introduces a minimal amount of prompt adherence. RenderDiffusion has higher prompt adherence but low quality, and our GANFusion can combine high quality with high prompt adherence due to our two-stage training strategy. Note that while the CLIP similarity is a cosine similarity and this has a maximum of $1$, we cannot reach this maximum, even with perfect prompt adherence, as text embeddings and image embeddings are separated in CLIP space~\cite{aggarwal2023Backdrop}. On the lower bound, we would also not expect the CLIP similarity to drop to $0$ without text conditioning, as all images in the dataset show humans and thus have some adherence to the prompt.

% \begin{figure*}[t]
%     \includegraphics[width=\textwidth]{images/qualitative_comparison_without_upsampling.pdf}
%     \caption{\textbf{Qualitative comparison without upsampling.}  We provide visual results from our method as well as the baselines. We disable the 2D upsampler for AG3D and our method and provide renderings of the triplane features directly for all methods.}
%     \label{fig:comparison_without_upsampling}
%     % \vspace{-1.4em}
% \end{figure*}

% \vspace{-1em}

\section{Conclusion, Limitations and Future Work}
In this work, we presented a novel text-to-3D generative model, which combines the strengths of GANs and Denoising Diffusion frameworks. Specifically, we demonstrate that the latent features learned with an unconditional GAN (EG3D or AG3D) using only 2D images as supervision can be exploited to train a text-guided denoising diffusion model. The resulting method allows to generate 3D geometry from text without test-time optimization by simply using standard diffusion in the learned triplane feature space. Overall, ours is the first text-guided feed-forward 3D generative model trained only with unstructured 2D data. %\maks{are we sure this is true?}

While our approach breaks new ground in efficient text-guided 3D synthesis, it comes with several limitations. First, any error in training the GAN will be inherited by the denoising diffusion network, which assumes a fixed latent space. Secondly, our current model works in a particular category, such as humans (or faces), and it would therefore be interesting to consider a \textit{general-purpose} text-guided 3D generator, capable of producing geometry across arbitrary classes or scenes. It will also be interesting to investigate other ways to condition the output, apart from text, using, e.g., directly providing depth or specific appearance, material or shape pose properties.
%While, in principle, possible with our framework, we have not investigated this possibility of using depth conditioning.
Finally, exploring other encodings, apart from the triplane features as done in our work, and using \textit{pre-trained} denoising diffusion networks, will potentially help to bridge the gap in terms of quality of output between the best image and 3D generative models.

\paragraph{Acknowledgements}
The authors would like to thank the anonymous reviewers for their valuable suggestions. Parts of this work were supported by the ERC Consolidfator Grant No. 101087347 (VEGA) and the ANR AI Chair AIGRETTE.

% \todo{mention that text conditioned ag3d figure, plus additional images (second half of figure 4) + supp will be in the appendix}

{\small
\bibliographystyle{ieee_fullname}
\bibliography{main}
}

\appendix
\newpage

\twocolumn[{%
 \centering
 {\Large \bf Supplementary Materials \par}

      % additional small space at the end of the author name
      \vskip .5em
      % additional empty line at the end of the title block
      \vspace*{12pt}
}]

% \begin{abstract}
In the following, we provide a detailed description of the BLIP~\cite{li2022blip}-based captioning process. Additionally, we include further qualitative examples and evaluate on additional datasets to illustrate the performance of our method in comparison to other baselines.

Alongside this document, the project webpage presents a qualitative comparison encompassing a broader set of randomly selected results from our method and all baselines. This webpage also includes videos for several examples.

% \end{abstract}

% \section{Additional Qualitative Results}
% Additional qualitative results are provided as webpage in the supplementary material.

% \section{Dataset Examples}

\section{Additional details for BLIP-based captioning}
\label{sec:blip_labeling}
We ask BLIP~\cite{li2022blip} the following questions:
\begin{itemize}
    \item What is the gender of this person? 
    \item What is this person wearing in top?
    \item What is the person wearing on their lower half?
    \item What color is this person wearing in top?
    \item What color is the clothing the person is wearing on their lower half?
    \item Is the person barefoot?
    \item What is the person wearing on their feet?
    \item What is the color of the thing the person is wearing on their feet?
\end{itemize}
and define a set of possible answers for each question:
\begin{itemize}
    \item \textbf{\texttt{Gender:}} 'man', 'woman'
    \item \textbf{\texttt{Color:}} 'Red', 'Blue', 'Green', 'Yellow', 'Purple', 'Orange', 'Black', 'White', 'Gray', 'Pink', 'Brown', 'Gold', 'Silver', 'Beige', 'Maroon', 'Teal', 'Olive', 'Navy', 'Coral', 'Turquoise', 'Indigo', 'Khaki'
    \item \textbf{\texttt{Upper body clothing:}} 'T-shirt', 'Polo shirt', 'Dress shirt', 'Tank top', 'Crop top', 'Blouse', 'Sweater', 'Hoodie', 'Jacket', 'Coat', 'Blazer', 'Vest', 'Sweatshirt', 'Pullover', 'Cardigan', 'Tunic', 'dress', 'Jumpsuit', 'Romper', 'Suit', 'Pajamas', 'Raincoats', 'Windbreakers', 'Parkas', 'Puffer jackets', 'Trench coats', 'Pea coats', 'Duffle coats'
    \item \textbf{\texttt{Lower body clothing:}} 'Jean', 'pant', 'Sweatpant', 'Legging', 'Short', 'Skirt', 'Capri', 'Chino', 'Jogger'
    \item \textbf{\texttt{Footwear:}} 'Sneakers', 'Loafers', 'boots', 'heels', 'Flats', 'Sandals', 'Flip flops', 'Espadrilles', 'Oxfords', 'Clogs', 'socks'
\end{itemize}
We use the answers chosen by BLIP to fill in the following prompt template:

\begin{itemize} [label={}]
    \item \texttt{[Prefix]} \texttt{[Gender]} wearing a \texttt{[Upper Body Color]} \texttt{[Upper body clothing]}, \texttt{[Lower body color]} \texttt{[Lower body clothing]}, \texttt{[Footwear color]} \texttt{[Footwear]}.
\end{itemize}
\texttt{[Prefix]} is randomly chosen from either \emph{``a photo of a''} or \emph{``a full-body photo of a''}, and \texttt{[Gender]} is selected randomly between the response of BLIP and \emph{``person.''}
At test time, we randomly drop some of the items in the prompt template, such as the upper body color and clothing, or the footwear color and clothing.

\begin{figure*}[t]
    \includegraphics[width=\textwidth]{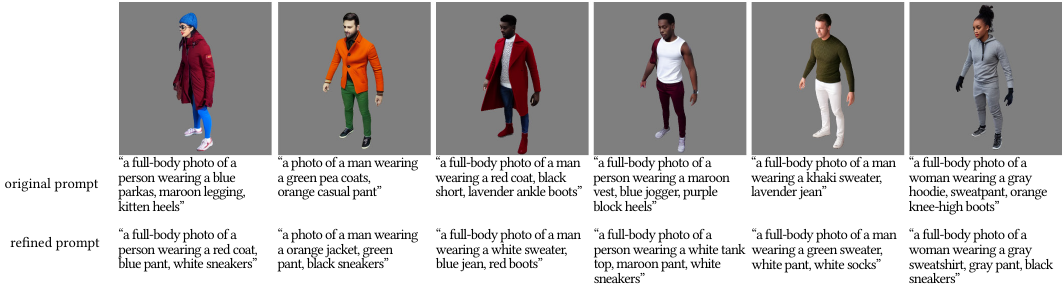}
    \caption{\textbf{Dataset examples.} We show a few images and corresponding prompts from our single-view image dataset. Prompts that were initially used to generate the images are often not accurate, we refine them with our BLIP-based captioning approach.}
    \label{fig:dataset}
    % \vspace{-3em}
\end{figure*}

\begin{figure*}[t!]
    \includegraphics[width=\textwidth]{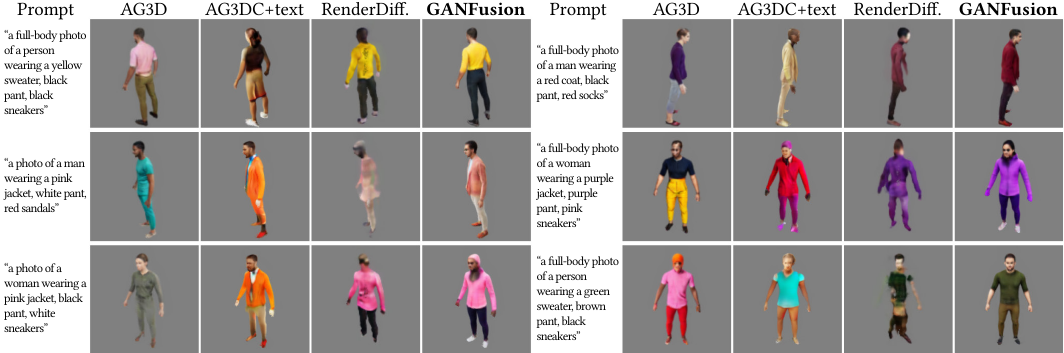}
    \caption{\textbf{Qualitative comparison without upsampling.}  We provide visual results from our method as well as the baselines. We disable the 2D upsampler for AG3D and our method and provide renderings of the triplane features directly for all methods.}
    \label{fig:comparison_without_upsampling}
    % \vspace{-1.4em}
\end{figure*}

\section{Additional Results}
\label{sec:add_results}

In this section, additional qualitative and quantitative results are provided.

In \cref{fig:ffhq}, we present qualitative results obtained by training our model on the FFHQ dataset\cite{chan2022eg3d}. We also evaluate our model quantitatively by measuring the FID of the generated samples and comparing our results to those of EG3D~\cite{chan2022eg3d}. Our model achieves an FID score of 49.4, compared to 26.7 achieved by EG3D. While our FID score is higher than that of EG3D, we emphasize that our model is conditioned on text, a capability that EG3D does not possess. We attribute the increased FID to the automatic labeling process used in stage 2. Specifically, the VQA model employed produces a limited set of labels, which restricts the variety of faces learned by our model compared to the full FFHQ dataset. This limitation could be addressed by utilizing a more powerful labeling algorithm.

\begin{figure*}[t]
    \includegraphics[width=\textwidth]{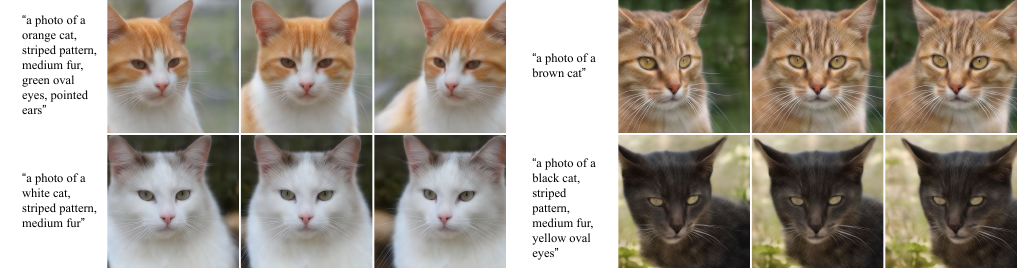}
    \caption{\textbf{Qualitative results on the AFHQ dataset~\cite{chan2022eg3d}.} We replace AG3D~\cite{dong2023ag3d} with EG3D~\cite{chan2022eg3d} as the generator in our first stage to effectively enable text-conditioning on real-world cat data.}
    \label{fig:afhq}
    % \vspace{-3em}
\end{figure*}

\begin{figure*}[t]
    \includegraphics[width=\textwidth]{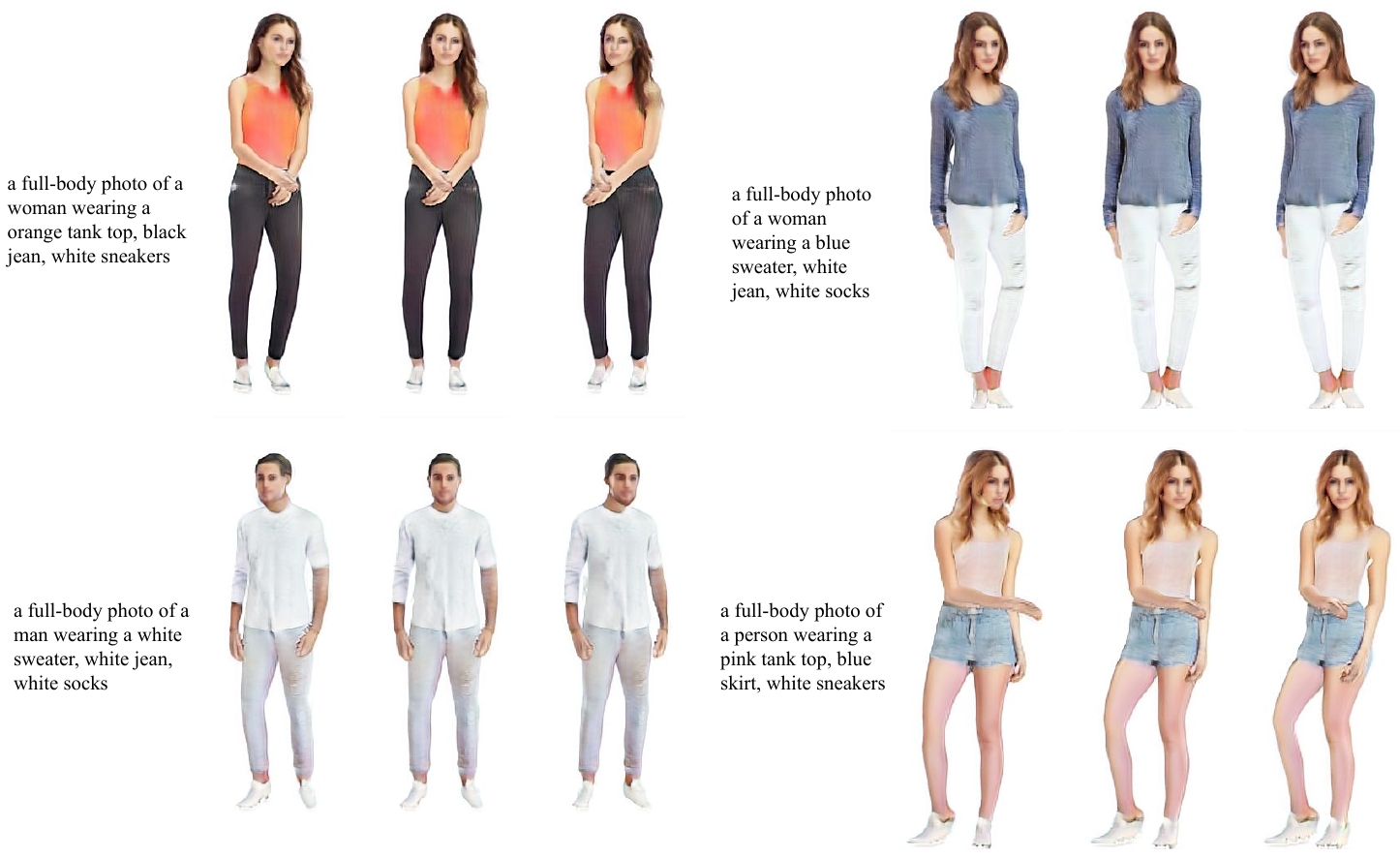}
    \caption{\textbf{Qualitative results on the DeepFashion dataset~\cite{deepfashion}.} We use AG3D~\cite{dong2023ag3d} as the generator in our first stage to effectively enable text-conditioning on realistic 3D human figures.}
    \label{fig:deepfashion}
    % \vspace{-3em}
\end{figure*}

Beyond learning human faces, we also trained our model to generate realistic cat faces from the AFHQ dataset~\cite{chan2022eg3d} using the EG3D~\cite{chan2022eg3d} backbone in stage 1, and realistic 3D human figures from the DeepFashion dataset~\cite{deepfashion} using the AG3D~\cite{dong2023ag3d} backbone in stage 1. Qualitative results for these experiments are provided in~\cref{fig:afhq} and~\cref{fig:deepfashion}, respectively. These results demonstrate that our model generates high-quality, realistic images while closely following the provided prompts. This highlights the adaptability of our model to multiple domains and its ability to leverage different backbones in stage 1.

\cref{fig:dataset} presents a selection of example images from the single-view image dataset we created, alongside their original captions and the refined captions, which were used to train our network.

Since RenderDiffusion does not employ an image upsampler, we also compared the results of the three main baselines without upsampling, as shown in~\cref{fig:comparison_without_upsampling}. Although omitting the upsampler reduces image detail in both AG3D and our method, both still significantly outperform RenderDiffusion.

\begin{figure*}[t]
    \includegraphics[width=\textwidth]{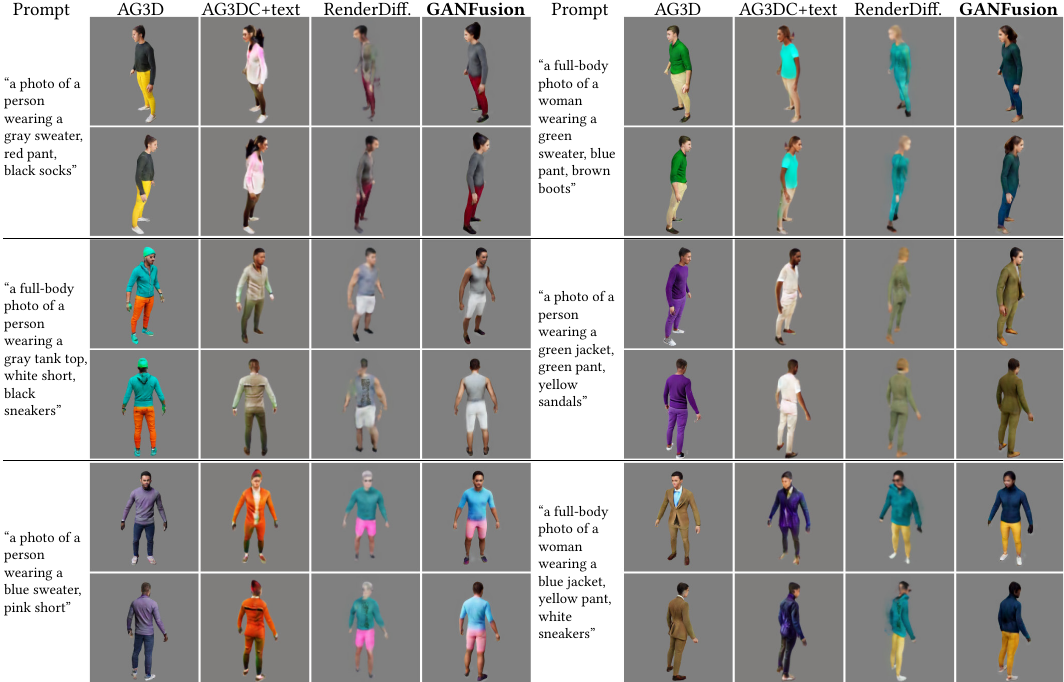}
    \caption{\textbf{Qualitative comparison.} We provide visual results from our method as well as the baselines.}
    \label{fig:comparison_supp}
    \vspace{-1.4em}
\end{figure*}

\begin{figure*}[t]
    \includegraphics[width=\textwidth]{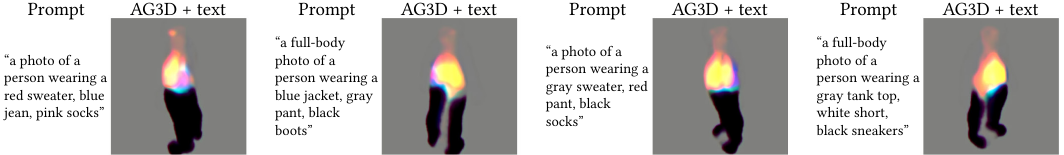}
    \caption{\textbf{Text-conditional AG3D.} We attempt to add text conditioning to AG3D~\cite{dong2023ag3d} by providing text embeddings as additional input both to the generator and the discriminator. However, we find that the training is not stable and does not converge.}
    \label{fig:conditional_ag3d}
\end{figure*}

\cref{fig:comparison_supp} showcases a qualitative comparison with all baselines, excluding the unsuccessful text-conditioned version of AG3D, which is presented in~\cref{fig:conditional_ag3d}. Due to training instability, experiments involving text-conditioning consistently led to mode collapse or other forms of training instabilities. The figure illustrates a case of severe mode collapse.

\begin{figure*}[t]
    \includegraphics[width=\textwidth]{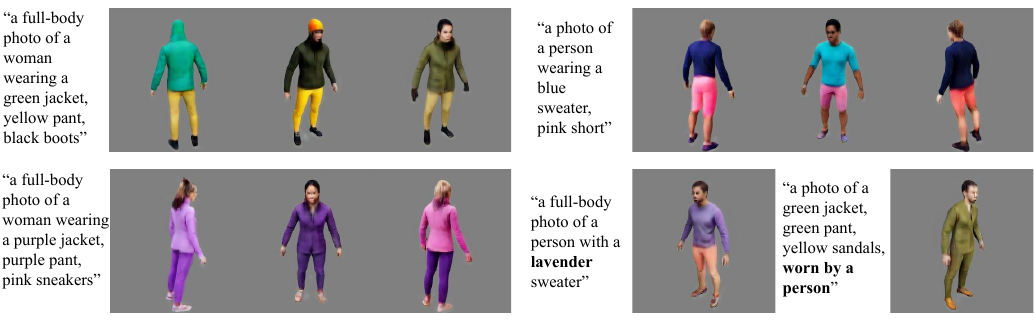}
    \caption{\textbf{Top \& Bottom left}: Diverse generations produced by our model using the same prompt but different seeds. \textbf{Bottom right}:  Images showcasing our model's ability to generalize to some out-of-distribution prompts not encountered during training.}
    \label{fig:diversity}
    % to modify: https://docs.google.com/presentation/d/1x9j80XLxaiDq59O_gelgMCfE53lnlHlGYuXmNhgBy7M/edit?usp=sharing
\end{figure*}

In~\cref{fig:diversity} (top row and bottom left), we showcase images produced by our method using identical prompts but different seeds, demonstrating the generation of diverse samples for the same prompt.

Furthermore, our primary objective is to demonstrate the feasibility of our approach using varied prompts that differ in properties such as colors and clothing items, rather than presenting a production-ready model. Generalizing to arbitrary prompts would require large-scale training on datasets like LAION \cite{schuhmann2022laion5b}, which is outside the scope of this study. Nonetheless, our model exhibits some generalization to out-of-distribution (OOD) prompts, facilitated by the pre-trained CLIP encoder. For instance, in~\cref{fig:diversity} (bottom right), our model correctly handles new colors like lavender, which are absent from the training dataset, and supports some degree of grammatical rearrangement in sentences.

% \begin{figure}[t]
%     \includegraphics[width=\columnwidth]{images/OOD.pdf}
%     \caption{\newtext{OOD generalization.}}
%     \label{fig:OOD}
%     % to modify: https://docs.google.com/presentation/d/1CCibhwJy1U6DGqsJZ3A_KoOnjuPlgckpf0Uwfr__O2Q/edit?usp=sharing
% \end{figure}

% \newtext{\paragraph{Acknowledgements}
% The authors would like to thank the anonymous reviewers for their valuable suggestions. Parts of this work were supported by the ERC Consolidfator Grant No. 101087347 (VEGA) and the ANR AI Chair AIGRETTE.}

% \todo{mention that text conditioned ag3d figure, plus additional images (second half of figure 4) + supp will be in the appendix}

% {\small
% \bibliographystyle{ieee_fullname}
% \bibliography{main}
% }

% \appendix

% \input{sections/figure_pages}
% \showthe\textwidth
% \showthe\columnwidth
% \showthe\linewidth

% \end{document}

\end{document}